\documentclass[a4paper]{spie}  

\usepackage{amsmath,amsfonts,amssymb}
\usepackage{bm}
\usepackage{graphicx}
\usepackage[colorlinks=true, allcolors=blue]{hyperref}
\usepackage{enumerate}
\usepackage{caption, subcaption}
\usepackage{booktabs}
\usepackage{threeparttable}

\title{iCBIR-Sli: Interpretable Content-Based Image Retrieval with 2D Slice Embeddings}

\author[a]{Shuhei Tomoshige}
\author[a]{Hayato Muraki}
\author[b]{Kenichi Oishi}
\author[a]{Hitoshi Iyatomi}
\affil[ ]{for the Alzheimer’s Disease Neuroimaging Initiative$^{*1}$}
\affil[ ]{the Australian Imaging Biomarkers and Lifestyle flagship study of ageing$^{*2}$}
\affil[a]{Dept. of Science and Engineering, Hosei University, 3-7-2 Kajino Koganei, Tokyo, 184-8584, Japan}
\affil[b]{Dept. of Radiology and Radiological Science, Johns Hopkins Medicine, 225 Traylor Building, 720 Rutland Ave.Baltimore, MD 21205, USA}

\authorinfo{$^{*1}$Data used in preparation of this article were obtained from the Alzheimer’s Disease Neuroimaging Initiative (ADNI) database (adni.loni.usc.edu). As such, the investigators within the ADNI contributed to the design and implementation of ADNI and/or provided data but did not participate in analysis or writing of this report. A complete listing of ADNI investigators can be found at: \url{http://adni.loni.usc.edu/wp-content/uploads/how_to_apply/ADNI_Acknowledgement_List.pdf}.\\$^{*2}$Data used in the preparation of this article was obtained from the Australian Imaging Biomarkers and Lifestyle flagship study of ageing (AIBL) funded by the Commonwealth Scientific and Industrial Research Organisation (CSIRO) which was made available at the ADNI database (\url{www.loni.usc.edu/ADNI}). The AIBL researchers contributed data but did not participate in analysis or writing of this report. AIBL researchers are listed at \url{www.aibl.csiro.au}.}

\begin{document} 
\maketitle

\begin{abstract}
Current methods for searching brain MR images rely on text-based approaches, highlighting a significant need for content-based image retrieval (CBIR) systems.
Directly applying 3D brain MR images to machine learning models offers the benefit of effectively learning the brain's structure; however, building the generalized model necessitates a large amount of training data.
While models that consider depth direction and utilize continuous 2D slices have demonstrated success in segmentation and classification tasks involving 3D data, concerns remain.
Specifically, using general 2D slices may lead to the oversight of pathological features and discontinuities in depth direction information.
Furthermore, to the best of the authors' knowledge, there have been no attempts to develop a practical CBIR system that preserves the entire brain's structural information.
In this study, we propose an interpretable CBIR method for brain MR images, named iCBIR-Sli (Interpretable CBIR with 2D Slice Embedding), which, for the first time globally, utilizes a series of 2D slices.
iCBIR-Sli addresses the challenges associated with using 2D slices by effectively aggregating slice information, thereby achieving low-dimensional representations with high completeness, usability, robustness, and interpretability—qualities essential for effective CBIR.
In retrieval evaluation experiments utilizing five publicly available brain MR datasets (ADNI2/3, OASIS3/4, AIBL) for Alzheimer's disease and cognitively normal, iCBIR-Sli demonstrated top-1 retrieval performance ($\text{macro F1}=0.859$), comparable to existing deep learning models explicitly designed for classification, without the need for an external classifier.
Additionally, the method provided high interpretability by clearly identifying the brain regions indicative of the searched-for disease.

\end{abstract}

\keywords{MRI, content-based image retrieval, 2D, 2.5D, Baseline++}

\section{INTRODUCTION}
\label{sec:intro}  

Brain magnetic resonance (MR) images, which can be obtained without radiation exposure, are frequently employed in the diagnosis of neurological disorders~\cite{morris1986nuclear}.
The acquired images are stored in the picture archive and communication system (PACS)~\cite{choplin1992picture}, where they are utilized for diagnostic support and research purposes~\cite{murala2013local}.
Typically, image queries within large-scale databases are performed through text-based retrieval.
However, this method relies heavily on the physician's expertise and experience to define appropriate labels and also necessitates an extensive number of annotations, leading to high costs.
Consequently, there is growing anticipation for developing content-based image retrieval (CBIR)~\cite{kumar2013content} techniques that enable the retrieval of similar cases based on image input. 
To realize practical CBIR, it is necessary to obtain a low-dimensional representation of brain MR images that possesses the following characteristics~\cite{cai2020content,nishimaki2022loc,muraki2024isometric}:
\begin{enumerate}[i.]
 \item Retention of brain structure and pathological features (completeness).\label{com}
 \item The ability to conduct searches using the representation itself (usability).\label{use}
 \item Continuity of data in the feature space (robustness).\label{robsut}
 \item Superior readability of the results derived (interpretability).\label{read}
\end{enumerate}

Several approaches utilizing machine learning have been proposed for acquiring low-dimensional representations of brain MR images~\cite{owais2019effective,swati2019content,arai2018significant,onga2019efficient,nishimaki2022loc}.
Muraki et al.~\cite{muraki2024isometric} introduced Isometric Feature Embedding for CBIR (IE-CBIR).
The low-dimensional representations generated by IE-CBIR are beneficial for CBIR as they facilitate disease retrieval purely based on the distances between these representations without needing external classifiers. 
In experiments involving Alzheimer's disease (AD) and cognitively normal (CN), IE-CBIR achieved disease detection capabilities comparable to the most advanced diagnostic methods, focusing solely on diagnosis.
Although this demonstrates that IE-CBIR possesses, to a certain extent, the aforementioned characteristics (\ref{com}), (\ref{use}), and (\ref{robsut}), there is room for further improvement regarding the interpretability of the representations.

In addition, while applying 3D machine learning models to brain MR images has the advantage of directly learning the brain's three-dimensional structure, concerns remain regarding the generalizability of these models. 
Although diverse training data is essential for 3D models, acquiring data in the medical field is highly costly. 
Moreover, 3D models are more complex than typical 2D models, leading to higher risks of overfitting and increased computational costs~\cite{zhang2022bridging,roy20222,emre2023pretrained}.
To overcome these challenges, it is necessary to enhance the model's generalization performance while efficiently using limited data. Consequently, in research targeting 3D brain images, the use of 2D or 2.5D models has been reported to yield superior results in common tasks such as classification and segmentation~\cite{liu2021style,swati2019content,zhang2022bridging,avesta2023comparing,roy20222,emre2023pretrained,nishimaki2024openmap}.
However, there are significant concerns regarding the use of 2D slice information for CBIR.
For instance, even if a patient has a particular disease, many slices may not exhibit pathological signs, and searching based solely on those slices would naturally yield inappropriate results.
To the best of our knowledge, no research has attempted to develop a practical CBIR system that retains the overall structural information of the brain while utilizing 2D slice images, nor has the feasibility of such a system been thoroughly examined.
Therefore, the objective of this research is to overcome the aforementioned concerns through appropriate processing, ultimately aiming to realize a more robust and interpretable practical CBIR system.

\section{Related work}
Kruthika et al.~\cite{kruthika2019cbir} proposed a CBIR (Content-Based Image Retrieval) system targeting Alzheimer's disease using 3D brain MR images and Capsule Networks~\cite{sabour2017dynamic}.
Unlike conventional CNNs, Capsule Networks enable the extraction of low-dimensional representations that take into account the relative structural information within the brain.
While these low-dimensional representations are robust against rotations and data transformations, their interpretability is not guaranteed.
Arai et al.~\cite{arai2018significant} attempted dimensionality reduction for brain MR images by employing a 3D extension of the convolutional autoencoder (CAE) to enable CBIR for brain MR images. They successfully obtained a 150-dimensional low-dimensional representation, reducing the number of elements from approximately 5 million per case to about 1/30,000 while retaining critical information necessary for disease identification and preserving brain structural details to a certain extent.
This study focused solely on the completeness requirement for CBIR, as mentioned earlier.
Subsequently, Onga et al.~\cite{onga2019efficient} proposed a method to derive low-dimensional representations by applying deep metric learning~\cite{oh2016deep} to the representations obtained by CAE. 
This method effectively reduced the influence of domain gaps caused by differences in data collection centers (domains), a significant issue in large-scale data analysis. 
It achieved this by responding more strongly to disease-specific features rather than individual differences, such as brain wrinkles, which are not inherently significant. 
This method utilizes the distances between the obtained low-dimensional representations, embodying the completeness criterion required for CBIR, where similar diseases share similar representations, while also offering the advantage of usability.
Nishimaki et al.~\cite{nishimaki2022loc} proposed a method called localized variational autoencoder (Loc-VAE) to acquire low-dimensional representations with enhanced interpretability for CBIR in brain MR images. Loc-VAE builds upon its backbone, the variational autoencoder (VAE)~\cite{DBLP:journals/corr/KingmaW13}, inheriting its feature of completeness. 
Additionally, it leverages the independence of each dimension in the low-dimensional representation and the robustness of the embedding derived from the continuity in the neighborhood of the data, thus addressing robustness. 
Furthermore, by introducing a novel loss function that restricts the information in each dimension to specific local regions in the original image, the method significantly improves the interpretability of the low-dimensional representations.

On the other hand, as an approach treating 3D data as a collection of 2D slices, Emre et al.~\cite{emre2023pretrained} proposed a two-stage model comprising a 2D CNN applied to individual slices of 3D data and an aggregation mechanism for the slice-level feature representations obtained by the CNN. 
Using retinal OCT data, they developed a model to predict the progression of age-related macular degeneration (AMD), reporting superior performance compared to 3D models.
Chen et al.~\cite{chen2023whole} proposed a whole-brain segmentation method using three consecutive 2D slices as input. 
This 2.5D approach, which processes bundles of 2D slices, has been widely adopted, particularly in segmentation tasks. 
It offers several advantages: reduced computational resources compared to 3D models, fewer training samples required, reduced risk of overfitting, and better preservation of inter-slice information compared to 2D methods.

In the broader field of machine learning, Chen et al.~\cite{DBLP:conf/iclr/ChenLKWH19} introduced a method called Baseline$++$, which utilizes prototypes~\cite{snell2017prototypical}—representative vectors for each class—and trains a model based on the cosine similarity between the obtained low-dimensional representations and these prototypes. 
This approach replaces the commonly used linear classifier in classification tasks with a distance-based classifier. 
By doing so, prototypes are optimized to maximize classification performance, forming clusters based on each class and minimizing intra-cluster variability. 
Data points within the same class are mapped to similar low-dimensional representations, enabling classification without the need for a separate classifier.
This makes the model highly compatible with CBIR systems.

\section{MATERIALS AND METHODS}
\subsection{Datasets}
For model training, we used the publicly available the Alzheimer's Disease Neuroimaging Initiative (ADNI) 2~\cite{mueller2005alzheimer,weiner2010alzheimer}. 
The ADNI was launched in 2003 as a public-private partnership, led by Principal Investigator Michael W. Weiner, MD. The primary goal of ADNI has been to test whether serial MRI, positron emission tomography (PET), other biological markers, and clinical and neuropsychological assessments can be combined to measure the progression of mild cognitive impairment (MCI) and early Alzheimer’s disease (AD). For up-to-date information, see \url{www.adni-info.org}. For this study, 1,884 CN cases and 1,003 AD cases from the ADNI2 dataset were used for training.

For model evaluation, we used data from ADNI2 that were not included in the training set, comprising 66 CN cases and 55 AD cases. To assess the model's generalizability, additional datasets were employed, including 352 CN cases and 91 AD cases from ADNI3, 452 CN cases and 260 AD cases from the Open Access Series of Imaging Studies (OASIS) 3/4~\cite{lamontagne2019oasis, koenig2020select}, and 235 CN cases and 45 AD cases from the Australian Imaging Biomarkers and Lifestyle (AIBL) dataset. AIBL study methodology has been reported previously~\cite{ellis2009australian}. For all datasets, a single image per patient was used during testing.

\subsection{iCBIR-Sli Overview}
\begin{figure}[t]
    \begin{minipage}[b]{0.38\columnwidth}
        \centering
        \includegraphics[width=\columnwidth]{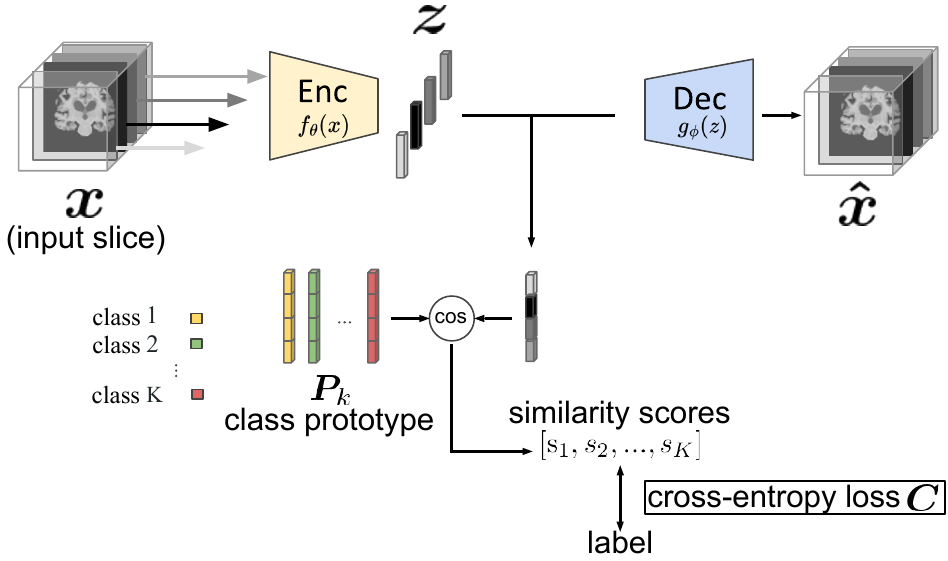}
        \subcaption{Architecture of iCBIR-Sli}
        \label{fig:2d_slice}
    \end{minipage}
    \hspace{0.02\columnwidth}
    \begin{minipage}[b]{0.59\columnwidth}
        \centering
        \includegraphics[width=\columnwidth]{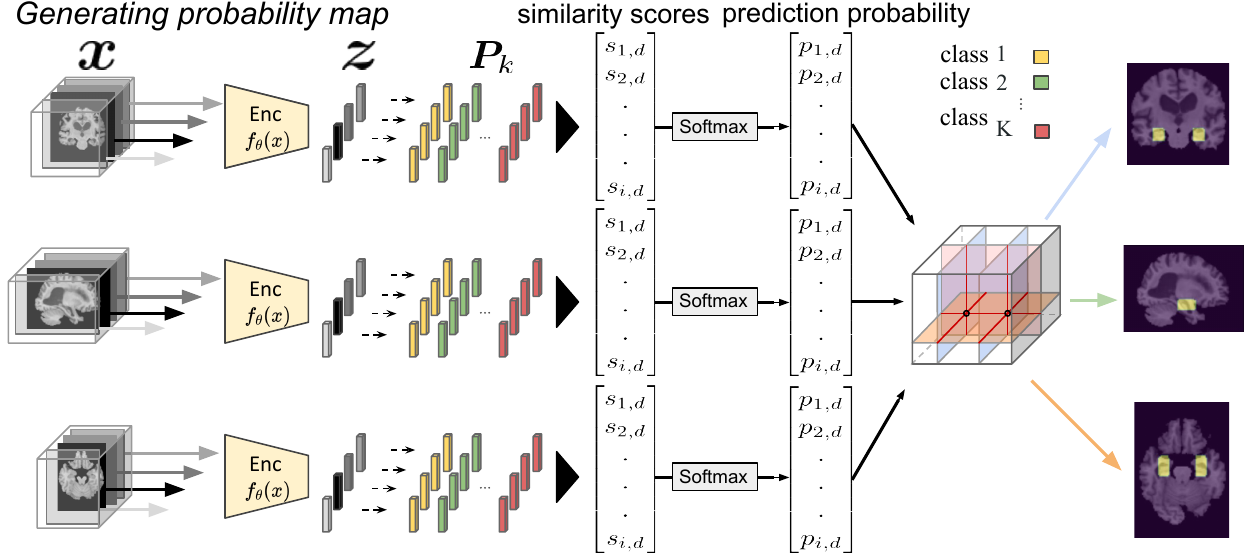}
        \subcaption{Generation of probability maps}
        \label{fig:proba}
    \end{minipage}
    \caption{Architecture of iCBIR-Sli and probability map generation for result interpretation.
}
    \label{fig:example}
\end{figure}

Fig.~\ref{fig:example} illustrates an overview of iCBIR-Sli.
In iCBIR-Sli, each 2D slice image $\bm{x}$ extracted from a 3D brain MR image is processed using VAE to obtain its low-dimensional representation $\bm{z}$.
Based on the set of obtained low-dimensional representations, our framework is performed by conducting distance-based nearest neighbor retrieval using Baseline$++$.
In other words, our framework extracts features using a 2D model and performs distance-based searches against disease prototypes obtained from Baseline$++$ using a so-called 2.5D representation, which is a concatenation of a certain number of low-dimensional representations from each slice.
This hybrid processing framework, combining 2D and 2.5D approaches, demonstrates excellent interpretability of the results. 
By calculating the distance between the low-dimensional representation of each slice and the corresponding slice of the class prototypes $\bm{P}_k$ ($k = 1, 2, \cdots, K$), the probability of each slice belonging to various diseases is determined using the Softmax function. 
These probabilities are then aggregated across three directions, enabling the generation of a probability map for each disease category at the 3D voxel level.

\subsection{iCBIR-Sli Implementation and Training}
The backbone of iCBIR-Sli, VAE, is a generative model that transforms each data point into a probability distribution. 
As a result, the obtained low-dimensional representation $\bm{z}$ preserves the brain MR image data while providing a continuous representation in that space. 
However, since VAE is an unsupervised learning method, the disease label information is not considered in the resulting low-dimensional representation $\bm{z}$, making it unsuitable for CBIR as is. 
The conventional approach of introducing a supervised learning classifier for similar case retrieval does not account for the interpretability of the low-dimensional representation $\bm{z}$, thereby reducing its usability for CBIR. 
To address this, iCBIR-Sli applies Baseline$++$, a method that replaces linear classifiers commonly used in the final stages of discriminative models with distance-based classifiers, to the low-dimensional representation $\bm{z}$.
With this approach, the low-dimensional representation $\bm{z}$ obtained by VAE from the input slice $\bm{x}$ is used to detect the disease class whose prototype $\bm{P}_k$ is closest in distance to $\bm{z}$, based on the principles of Baseline$++$. 
As with Baseline$++$, the initial value of the prototype $\bm{P}_k$ for each disease class is set as the average value of $\bm{z}$ belonging to that class.
In this context, the vector $\bm{z}$ and the prototypes $\bm{P}_k$ for the $k$ classes are normalized to have a magnitude of 1, and the similarity $s_k$ of $\bm{z}$ to class $k$ is calculated using cosine similarity $R$:
\begin{equation}
    \label{eq:cos}
    s_k = R(\frac{\bm{z}}{||\bm{z}||}, \frac{\bm{P}_{k}}{||\bm{P}_{k}||})
\end{equation}
The estimated disease class $s^*$ for this slice is determined as the class with the highest $s_k$:
\begin{equation}
    \label{eq:s^}
    s^* = \operatorname{argmax}(s_k)
\end{equation}
Note that this process is the same when processing multiple slices that will be used later.
Next, we introduce an error term for training the prototypes $\bm{P}_k$.
The predicted probability vector $\bm{p}$ of each disease class for the input $\bm{x}$ can be expressed as follows, using $\mathcal{S} = [s_1, s_2, \cdots, s_k, \cdots, s_K]^\top$:
\begin{equation}
    \label{eq:p}
    \bm{p} = Softmax (\mathcal{S})
\end{equation}
The difference between this value and the corresponding disease class label $\bm{t}$ (a one-hot vector) is computed as the cross-entropy loss $\mathcal{C}$, which is then used to train each prototype $\bm{P}_k$ and the encoder to construct a more refined low-dimensional representation:
\begin{equation}
    \label{eq:ce}
    \mathcal{C}=CrossEntropy (\bm{p},\bm{t})
\end{equation}
As a result, the low-dimensional representation $\bm{z}$ is learned to approximate the prototype $\bm{P}_{k=d}$ of the corresponding disease class $d$ while preserving the original brain structural information.
This allows for disease retrieval based solely on the distances between the low-dimensional representations $\bm{z}$, without the need for an external classifier, offering a significant advantage for the realization of CBIR.
Finally, the loss function of iCBIR-Sli is as follows:
\begin{equation}
    \label{eq:Sli}
    \mathcal{L}=\mathcal{D}(\bm{x},\bm{\hat{x}})+\beta\mathcal{D}_{KL}[q(\bm{z}|\bm{x})||p(\bm{z})]+\gamma\mathcal{C},
\end{equation}
where $\mathcal{D}$ is the reconstruction error, for which the mean squared error (MSE) was used in this experiment, and $\mathcal{D}_{KL}$ is the Kullback-Leibler (KL) divergence between the estimated generative distribution $q(\bm{z}|\bm{x})$ obtained by the encoder and the prior distribution $p(\bm{z})$, which is assumed to be a normal distribution for $\bm{z}$. 
These terms constitute the loss function of VAE, with $\mathcal{D}$ encouraging accurate reconstruction of the input and $\mathcal{D}_{KL}$ regularizing the latent space. The $\beta$ and $\gamma$ are hyperparameters, respectively.
Based on this loss, the training of the encoder, decoder, and the disease prototypes $\bm{P}_k$ used in Baseline$++$ is conducted.

\subsection{Similar case retrieval using iCBIR-Sli}
As previously mentioned, since each slice may often lack the necessary information, iCBIR-Sli performs searches using consecutive slices.
Specifically, the low-dimensional representations $\bm{z}_i$ ($i = 1, 2, \cdots, N_{\mathrm{section}}$) of each slice obtained from VAE are grouped into blocks of $n$ slices each, and $m$ slices slide these blocks to construct a total of $J$ box-shaped regions across each cross-section.
The low-dimensional representation of $j$-th block, $\bm{b}_j$ ($j = 1, 2, \cdots, j, \cdots, J$) and the corresponding $k$-class prototype at that position, $\bm{u}_{j,k}$, are simply the concatenations of the low-dimensional representations and prototypes of the respective 2D slices.
Within each block region, the disease most closely matching the prototype is identified following the same procedure as outlined in equations (\ref{eq:cos}) - (\ref{eq:p}).
\begin{equation}
    \label{eq:block}
    \bm{b}_j = [\bm{z}_{(j-1)m+1}, \bm{z}_{(j-1)m+2}, \cdots, \bm{z}_{(j-1)m+n}]^\top
\end{equation}
This process is repeated for all blocks, and if the number of diagnoses for a given disease $d$ exceeds the predetermined threshold value $\xi_d$, the input image is classified as having disease $d$.
Additionally, if this process is applied to each cross-section and pathological features are detected in at least the pre-specified number of cross-sections $r$ ($r = 1, 2, 3$), the input image is deemed to exhibit those pathological features.
In this experiment, we adopted $r = 1$.

Next, we describe the generation of probability maps for each disease, with an overview provided in Fig.~\ref{fig:proba}. As illustrated, the process of generating disease probability maps differs from the block processing used in similar case retrieval methods. The probability of each disease class for each slice is calculated using equations (\ref{eq:cos}) – (\ref{eq:p}), and these probabilities are aggregated across cross-sections. Subsequently, the results from the three cross-sections are integrated to generate the probability map for each disease at the 3D voxel level.
In other words, this approach provides a high level of interpretability (\ref{read}) for the results of CBIR.
In practical CBIR applications, the probability map for a disease is output when the disease is present in the search results for a given cross-section, thereby providing a rationale for the search results.

\section{Results and Discussion}
\label{sec:sections}

\begin{table*}[t]
    \caption{Classification (neighborhood search) performance for each model.}
    \label{tab:result}
    \begin{minipage}{\linewidth}
        \centering
        \resizebox{\textwidth}{!}{
        \begin{tabular}{@{}lrrrrrrr@{}}
            \toprule
            & \multicolumn{3}{c}{\textbf{ADNI2~\cite{mueller2005alzheimer,weiner2010alzheimer}}} & \multicolumn{3}{c}{\textbf{ADNI3~\cite{mueller2005alzheimer,weiner2010alzheimer}}} \\ \cmidrule(lr){2-4} \cmidrule(lr){5-7}
            & \multicolumn{1}{c}{\textbf{Precision}} & \multicolumn{1}{c}{\textbf{Recall}} & \multicolumn{1}{c}{\textbf{F1}} & \multicolumn{1}{c}{\textbf{Precision}} & \multicolumn{1}{c}{\textbf{Recall}} & \multicolumn{1}{c}{\textbf{F1}} & \\ \cmidrule(r){1-1} \cmidrule(lr){2-2} \cmidrule(lr){3-3} \cmidrule(lr){4-4} \cmidrule(lr){5-5} \cmidrule(lr){6-6} \cmidrule(lr){7-7}
            3D-CNN & 0.818 & 0.818 & 0.820 & 0.802 & 0.758 & 0.862 \\
            IE-CBIR~\cite{muraki2024isometric} & 0.914 & 0.762 & 0.873 & 0.812 & 0.793 & 0.878 \\
            \cmidrule(r){1-1}\cmidrule(lr){2-2} \cmidrule(lr){3-3} \cmidrule(lr){4-4} \cmidrule(lr){5-5} \cmidrule(lr){6-6}\cmidrule(lr){7-7}
            \textbf{iCBIR-Sli$^\dagger$} & 0.905/0.909/0.844 & 0.760/0.600/0.760 & 0.856/0.784/0.831 & 0.838/0.847/0.826 & 0.703/0.586/0.729 & 0.856/0.784/0.831 \\
            \textbf{ensemble} & 0.840 & 0.840 & 0.859 & 0.777 & 0.791 & 0.864 \\ \bottomrule
        \end{tabular}}
    \end{minipage}
    \vspace{2mm}
    \begin{minipage}{\linewidth}
        \centering
        \resizebox{\textwidth}{!}{
        \begin{tabular}{@{}lrrrrrrr@{}}
            & \multicolumn{3}{c}{\textbf{OASIS~\cite{lamontagne2019oasis, koenig2020select}}} & \multicolumn{3}{c}{\textbf{AIBL~\cite{ellis2009australian}}} \\ \cmidrule(lr){2-4} \cmidrule(lr){5-7}
            & \multicolumn{1}{c}{\textbf{Precision}} & \multicolumn{1}{c}{\textbf{Recall}} & \multicolumn{1}{c}{\textbf{F1}} & \multicolumn{1}{c}{\textbf{Precision}} & \multicolumn{1}{c}{\textbf{Recall}} & \multicolumn{1}{c}{\textbf{F1}} & \\ \cmidrule(r){1-1} \cmidrule(lr){2-2} \cmidrule(lr){3-3} \cmidrule(lr){4-4} \cmidrule(lr){5-5} \cmidrule(lr){6-6} \cmidrule(lr){7-7}
            3D-CNN & 0.802 & 0.762 & 0.830 & 0.440 & 0.841 & 0.726 \\
            IE-CBIR~\cite{muraki2024isometric} & 0.848 & 0.669 & 0.813 & 0.514 & 0.841 & 0.771 \\
            \cmidrule(r){1-1}\cmidrule(lr){2-2} \cmidrule(lr){3-3} \cmidrule(lr){4-4} \cmidrule(lr){5-5} \cmidrule(lr){6-6}\cmidrule(lr){7-7}
            \textbf{iCBIR-Sli$^\dagger$} & 0.817/0.888/0.868 & 0.655/0.568/0.685 & 0.796/0.781/0.826 & 0.402/0.654/0.454 & 0.818/0.758/0.871 & 0.697/0.820/0.737 \\
            \textbf{ensemble} & 0.787 & 0.786 & 0.832 & 0.356 & 0.909 & 0.661 \\ \bottomrule
        \end{tabular}}
        \begin{tablenotes}
            \footnotesize
            \item $\dagger$: coronal/sagittal/axial cross-section. \textbf{Unlike other approaches, iCBIR-Sli has the unique ability to visually present the basis for diagnosis.}
        \end{tablenotes}
    \end{minipage}
\end{table*}
\begin{figure} [t]
   \begin{center}
   \includegraphics[height=3.4cm, keepaspectratio]{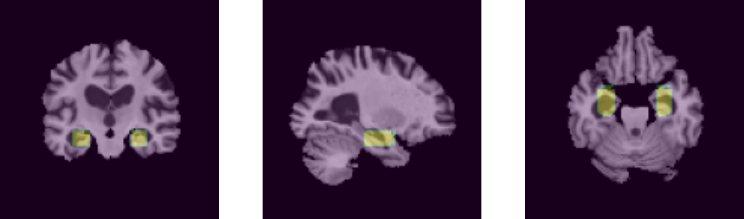}
   \end{center}
   \caption[example] 
   { \label{fig:prob}
   Visualization of probability maps at each cross-section ($\xi_d$ = 0.8). The yellow highlights indicate the areas that have been visualized as the basis for the diagnosis.}
\end{figure} 
We conducted a retrieval evaluation experiment for AD using the iCBIR-Sli framework with ADNI2/3~\cite{mueller2005alzheimer,weiner2010alzheimer}, OASIS~\cite{lamontagne2019oasis, koenig2020select}, and AIBL~\cite{ellis2009australian}.
For comparative analysis, we employed two methods: a 3D-CNN model exclusively specialized for disease diagnosis and IE-CBIR, a state-of-the-art CBIR model specifically designed for superior retrieval performance.
Tab.~\ref{tab:result} presents a comparison of the retrieval results of our proposed and comparative methods.

In ADNI2, iCBIR-Sli achieved a top-1 retrieval performance ($\text{macro F1}=0.859$) comparable to that of existing deep learning models specialized in classification based solely on the learned low-dimensional representations. 
In ADNI3 dataset, the proposed method achieved performance comparable to that of CNN.
For OASIS dataset, it outperformed both CNN and IE-CBIR.
Regarding AIBL dataset, although the overall performance in the ensemble setting decreased due to the recall-focused configuration with
$r = 1$ and the limited number of AD cases, the sagittal view achieved superior performance compared to other methods.
In addition, both IE-CBIR and iCBIR-Sli possess the advantage of enabling disease retrieval solely through the obtained low-dimensional representations.
However, the proposed iCBIR-Sli surpasses IE-CBIR by providing a clear localization of the brain regions that serve as evidence for the queried image. 
Fig.~\ref{fig:prob} illustrates the probability maps generated by iCBIR-Sli for correctly retrieved AD cases, highlighting regions with an 80\% or higher probability of being associated with AD and overlaying them on the input image.
The highlighted regions correspond to the hippocampus, aligning with well-established medical knowledge.
This demonstrates that our model can visually present the specific areas it emphasizes during inference, providing an interpretable output.
This result indicates that a practical CBIR system that retains comprehensive structural information of the brain, previously unattempted, is achievable by effectively utilizing 2D slice images.

\section{Conclusions}
In this paper, we proposed the iCBIR-Sli framework to realize CBIR for brain MR images.
To achieve practical CBIR, it is essential to obtain low-dimensional representations that satisfy the following criteria: completeness, usability, robustness, and interpretability.
Our experiment demonstrated that iCBIR-Sli achieved performance comparable to or exceeding that of existing 3D techniques, successfully obtaining low-dimensional representations that meet the aforementioned requirements.
Notably, the unprecedented interpretability of iCBIR-Sli lies in its ability to provide voxel-level probability maps that pinpoint the specific regions deemed.
Although this study only involved two classes (AD and CN), we believe that iCBIR-Sli can be adapted to a wide variety of diseases beyond specific conditions.
Future work will include validating its performance when additional disease categories are incorporated.

\section{ACKNOWLEDGMENTS}
This research was supported in part by the Ministry of Education, Science, Sports and Culture of Japan (JSPS KAKENHI), Grant-in-Aid for Scientific Research (C), 21K12656 (2021-2024) and 24K17506 (2024-2027). \\
Data collection and sharing for this project was funded by the Alzheimer's Disease Neuroimaging Initiative
(ADNI) (National Institutes of Health Grant U01 AG024904) and DOD ADNI (Department of Defense award
number W81XWH-12-2-0012). ADNI is funded by the National Institute on Aging, the National Institute of
Biomedical Imaging and Bioengineering, and through generous contributions from the following: AbbVie,
Alzheimer’s Association; Alzheimer’s Drug Discovery Foundation; Araclon Biotech; BioClinica, Inc.; Biogen;
Bristol-Myers Squibb Company; CereSpir, Inc.; Cogstate; Eisai Inc.; Elan Pharmaceuticals, Inc.; Eli Lilly and
Company; EuroImmun; F. Hoffmann-La Roche Ltd and its affiliated company Genentech, Inc.; Fujirebio; GE
Healthcare; IXICO Ltd.; Janssen Alzheimer Immunotherapy Research \& Development, LLC.; Johnson \&
Johnson Pharmaceutical Research \& Development LLC.; Lumosity; Lundbeck; Merck \& Co., Inc.; Meso
Scale Diagnostics, LLC.; NeuroRx Research; Neurotrack Technologies; Novartis Pharmaceuticals
Corporation; Pfizer Inc.; Piramal Imaging; Servier; Takeda Pharmaceutical Company; and Transition
Therapeutics. The Canadian Institutes of Health Research is providing funds to support ADNI clinical sites
in Canada. Private sector contributions are facilitated by the Foundation for the National Institutes of Health
(\url{www.fnih.org}). The grantee organization is the Northern California Institute for Research and Education,
and the study is coordinated by the Alzheimer’s Therapeutic Research Institute at the University of Southern
California. ADNI data are disseminated by the Laboratory for Neuro Imaging at the University of Southern
California.\\
Data were provided by OASIS3/4 Longitudinal Multimodal Neuroimaging: Principal Investigators: T. Benzinger, D. Marcus, J. Morris; NIH P30 AG066444, P50 AG00561, P30 NS09857781, P01 AG026276, P01 AG003991, R01 AG043434, UL1 TR000448, R01 EB009352. AV-45 doses were provided by Avid Radiopharmaceuticals, a wholly owned subsidiary of Eli Lilly, and Clinical Cohort: Principal Investigators: T. Benzinger, L. Koenig, P. LaMontagne.

\bibliography{ref_tomoshi} 
\bibliographystyle{spiebib}

\end{document}